\documentclass[a4paper,12pt]{article}

\usepackage{comment}
\usepackage{fancyvrb}
\usepackage{graphicx}
\usepackage{hyperref}
\usepackage[ragged]{footmisc}
\usepackage{ragged2e}
\usepackage[section]{placeins}
\usepackage{geometry}
\usepackage{caption}

\begin{document}

\title{Transparency and granularity in the SP Theory of Intelligence and its realisation in the SP Computer Model\footnote{Published in the book {\em Interpretable Artificial Intelligence: A Perspective of Granular Computing}, Witold Pedrycz and Shyi-Ming Chen (editors), Springer: Heidelberg, 2021, ISBN 978-3-030-64948-7, DOI: 10.1007/978-3-030-64949-4.}}

\author{J Gerard Wolff$^1$ \\
\\
$^1$CognitionResearch.org, Menai Bridge, UK \\
\href{mailto:jgw@cognitionresearch.org}{jgw@cognitionresearch.org}}

\date{}

\maketitle

\begin{abstract}

This chapter describes how the {\em SP System}, meaning the {\em SP Theory of Intelligence}, and its realisation as the {\em SP Computer Model}, may promote transparency and granularity in AI, and some other areas of application. The chapter describes how transparency in the workings and output of the SP Computer Model may be achieved via three routes: 1) the program provides a very full audit trail for such processes as recognition, reasoning, analysis of language, and so on. There is also an explicit audit trail for the unsupervised learning of new knowledge; 2) knowledge from the system is likely to be granular and easy for people to understand; and 3) there are seven principles for the organisation of knowledge which are central in the workings of the SP System and also very familiar to people (eg chunking-with-codes, part-whole hierarchies, and class-inclusion hierarchies), and that kind of familiarity in the way knowledge is structured by the system, is likely to be important in the interpretability, explainability, and transparency of that knowledge. Examples from the SP Computer Model are shown throughout the chapter.

\end{abstract}


\captionsetup[figure]{labelfont={bf},labelformat={default},labelsep=period,name={Fig.}}

\section{Introduction}\label{introduction-section}

As its title suggests, the subject of this chapter is how the {\em SP System}, which means the {\em SP Theory of Intelligence}, and its realisation as the {\em SP Computer Model}, may promote `transparency' and `granularity' in both the structuring and processing of knowledge. It is chiefly relevant to AI and the modelling of human learning, perception, and cognition (HLPC) but, as will be described later, it is also relevant to other areas of application for the SP System.

Since people often ask about the meaning of `SP', it is intended to be a name, not an abbreviation. That said, Section \ref{ic-in-sp-system-section} describes the meanings of `S' and `P' in the SP programme of research.

The next two sections contain introductory remarks about transparency and granularity. Then Section \ref{sp-outline-section} outlines the SP System with pointers to where fuller information may be found. Section \ref{ic-kr-pr-in-sp-system-section} discusses information compression (IC) and the representation and processing of knowledge in the SP System. Section \ref{transparency-via-audit-trails-section} describes how the SP System promotes transparency via the provision of audit trails for all its processing. Section \ref{transparency-granularity-familiarity-section} discusses how granularity and corresponding transparency may be seen in the workings of the SP System. And, since the knowledge created by the SP System will be in forms such as chunking-with-codes, part-whole-hierarchies, and class-inclusion hierchies that are widely used by people, that very familiarity will facilitate interpretability, explainability, and consequent transparency in that knowledge. Section \ref{interpretability-explainability-section} outlines some recent studies related to interpretability and explainability, with some brief comments.

\section{Introduction to transparency}\label{transparency-introduction-section}

In the words of the `call for chapters' for this book: ``It is desirable that the models of AI are {\em transparent} so that the results being produced have to be easily {\em interpretable} and {\em explainable}.'' (emphasis added). Thus transparency' in an AI program means that processing by the program and results from it are comprehensible by people, and thus interpretable and explainable. Hence, the main emphasis in this chapter is on transparency rather than the more specific concepts of interpretable and explainable, but see Section \ref{interpretability-explainability-section}.

Transparency in AI systems is a matter of concern, chiefly because of shortcomings in deep neural networks (DNNs). Despite their striking successes in several different areas of application, it is normally difficult to understand how DNN results are achieved.

Transparency is particularly important when there is a need to diagnose what has gone wrong when there are outright failures of DNNs, and these can be dramatic. For example, a DNN may fail to recognise something in a picture that, to a person, looks very similar to another picture which the DNN recognises correctly \cite{szegedy-etal-2014}. And a DNN may recognise something as a `guitar' or a `penguin', when people think it looks like white noise on a TV screen or an abstract pattern that does not represent any specific object \cite{nguyen-etal-2015}.

Failures of DNNs can be can be both expensive and dangerous, either or both of which may apply if, for example, self-driving cars were too dependent on DNNs for the recognition of objects and events as they drive along. Obscurity in the workings of DNNs makes it difficult to find out the reason or reasons for such failures.

\section{Introduction to granularity}\label{granularity-introduction-section}

The concept of an `information granule' has a bearing on transparency in computing and thus the interpretablity of the results of computing, and their explainability. The concept has been defined as:

\begin{quote}

    ``...~a clump of points (objects) drawn together by indistinguishability, similarity, proximity or functionality. For example, the granules of a human head are the forehead, nose, cheeks, ears, eyes, etc. In general, granulation is hierarchical in nature. In general, granulation is hierarchical in nature. A familiar example is the
    granulation of time into years, months, days, hours, minutes, etc.'' \cite[p.~111]{zadeh-1997}.

\end{quote}

\noindent And in a similar way, an `information granule' is:

\begin{quote}

    ``...~a collection of elements drawn together by their closeness (resemblance, proximity, functionality, etc.) articulated in terms of some useful spatial, temporal, or functional relationship. ...~Granular Computing is about representing, constructing, processing, and communicating information granules.'' \cite[p.~1026]{pedrycz-2018}.

\end{quote}

\section{The SP System in brief}\label{sp-outline-section}

The SP Theory of Intelligence with the SP Computer Model are the products of {\em a lengthy research programme which has aimed to simplify and integrate observations and concepts across AI, mainstream computing, mathematics, and HLPC}.
Despite its ambition, that goal has been largely met, with corresponding benefits in the versatility of the SP System, as outlined in Sections \ref{strengths-potential-section} and \ref{non-ai-areas-of-application-section}, below.

\sloppy This section aims provide readers with enough information to make the rest of the chapter intelligible. There is more information in \cite{sp-extended-overview}, which is largely a shortened version of the book {\em Unifying Computing and Cognition} \cite{wolff-2006}, in which much fuller information may be found. Details of other peer-reviewed papers, and other documents, with download links, are in \href{http://www.cognitionresearch.org/sp.htm}{www.cognitionresearch.org/sp.htm}.

Readers may wonder why most of the citations of research related to the SP System are to this author's work. The main reasons seem to be these: 1) The SP System is radically different from DNNs which dominate AI research today, and it is well known that radical ideas can take time to gain acceptance \cite{kuhn-2012}; 2) When the book {\em Unifying Computing and Cognition} was published in 2006, the author switched his attention into nearly seven years of full-time campaigning about climate change, instead of promoting the book as would normally have been the case; 3) Because, for environmental reasons, flying has been avoided as much as possible, it has not been possible to attend many of the influential conferences such as IJCAI; 4) Given the intense pressures on many researchers to ``publish or perish'', it can be difficult for them to get into a new field.

\subsection{Information compression}\label{ic-in-sp-system-section}

Right from the beginning of this research, a unifying theme, which has proved its value in spades, is that IC in the SP System is likely to be part of the solution to the goal of simplification and integration across a broad canvass. This is largely because of an accumulation of evidence from many studies, beginning with pioneering research by Fred Attneave \cite{attneave-1954} and Horace Barlow \cite{barlow-1959,barlow-1969}, and others, showing the importance of IC in HLPC \cite{sp-compression}.

The letters `S' and `P' in the name `SP' may be seen to stand for `Simplicity' and `Power'. This is because: 1) a good theory should combine conceptual `Simplicity' with explanatory or descriptive `Power'; and 2) IC, which is central in the organisation and workings of the SP System, may be seen as a process which increases the `Simplicity' of a body of information, {\bf I}, by the extraction of unnecessary repetition or redundancy in {\bf I}, and at the same time retains as much as possible of its explanatory and descriptive `Power'.

There is more detail about IC in the SP System in Sections \ref{concept-of-spma-section}, \ref{unsupervised-learning-section}, and \ref{ic-kr-pr-in-sp-system-section}.

\subsection{Abstract view of the SP System}\label{sp-abstract-section}

\begin{figure}[!htbp]
\centering
\includegraphics[width=0.5\textwidth]{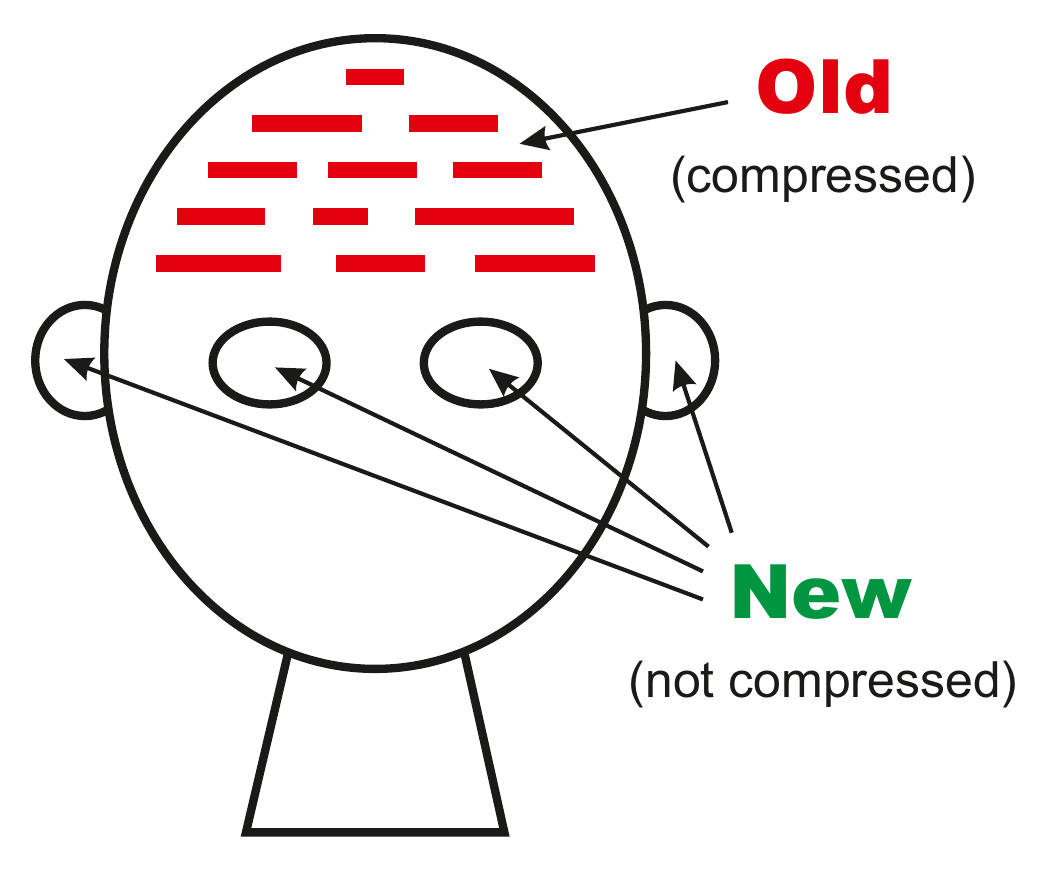}
\caption{Schematic representation of the SP System. Adapted from Figure 1 in \cite{sp-extended-overview}, with permission.}
\label{sp-input-perspective-figure}
\end{figure}

At a high level of abstraction, the SP System may be seen to be like a brain which takes in {\em New} information (with a capital `N') through its senses and stores all or part of it in a repository of {\em Old} information (with a capital `O'), as shown schematically in Figure \ref{sp-input-perspective-figure}.

\subsection{Basic structures in the SP System for representing knowledge}\label{sp-patterns-and-symbols-section}

In the SP System, {\em SP-patterns} are the vehicle for storing all kinds knowledge. Here an SP-pattern is an array of atomic {\em SP-symbols} in one or two dimensions. The SP Computer Model has not yet been developed to process two-dimensional SP-patterns, but the aim is for that to be possible in later versions of the program.

An SP-symbol is simply a mark that can be matched with any other symbol to yield a `same' or `different' answer. The meaning of any SP-symbol is provided exclusively by its association with one or more other SP-symbols. There is nothing like $+$ or $\times$ in arithmetic, where the meaning of each of those two symbols is hidden from view. As described in Section \ref{concept-of-spma-section}, below, SP-symbols gain expressive power via their roles in SP-multiple-alignments.

An SP-pattern that is `New' is raw information from the system's `environment', brought in via its `senses'. An example of such a New SP-pattern is the sentence `\texttt{f o r t u n e f a v o u r s t h e b r a v e}' in row 0 in Figure \ref{fortune-brave-spma-figure}, below.

Each SP-pattern that is Old has `ID' SP-symbols at the beginning and end which are used in the building of `SP-multiple-alignments' and in the encoding process, as outlined in Section \ref{concept-of-spma-section}, below.

An example of an Old SP-pattern as just described is the SP-pattern `\texttt{N 4 f o r t u n e \#N}' in row 4 in Figure \ref{fortune-brave-spma-figure}, below. Here, the SP-symbols `\texttt{N}', `\texttt{4}', and `\texttt{\#N}' are examples of the `ID' SP-symbols mentioned above. The ID SP-symbols `\texttt{N}' and `\texttt{\#N}' serve in marking the start and finish of the SP-pattern and also in classifying the SP-pattern as a noun. The ID SP-symbol `\texttt{4}' distinguishes this noun from others in the set of stored Old SP-patterns.

Although SP-patterns and SP-symbols are very simple, they gain expressive power via their roles in SP-multiple-alignments, (see, Section \ref{concept-of-spma-section}, next).

Provided that SP-patterns have been created via unsupervised learning that achieves high levels of IC (Section \ref{unsupervised-learning-section}), it seems likely that they would be amongst the groupings recognised as `granules', and also `chunks' (Section \ref{miller-chunk-section}) and `objects' or `entities' (Section \ref{oo-section}).

\subsection{The concept of SP-multiple-alignment}\label{concept-of-spma-section}

A central part of the SP Computer Model, is the concept of {\em SP-multiple-alignment} (SPMA). It is a concept that has been adapted from the concept of `multiple sequence alignment' in bioinformatics.

This concept is responsible for most of the existing and potential versatility of the SP System in all areas of AI except unsupervised learning, but even in unsupervised learning it has a major role to play. Existing and potential strengths of the SP System are summarised in Sections \ref{strengths-potential-section} and \ref{non-ai-areas-of-application-section}.

\subsubsection{Multiple sequence alignment}

As an introduction to the concept of SPMA, Figure \ref{ma-dna-figure} shows an example of a multiple sequence alignment.

\begin{figure}[!htbp]
\fontsize{10.00pt}{12.00pt}
\centering
{\bf
\begin{BVerbatim}
  G G A     G     C A G G G A G G A     T G     G   G G A
  | | |     |     | | | | | | | | |     | |     |   | | |
  G G | G   G C C C A G G G A G G A     | G G C G   G G A
  | | |     | | | | | | | | | | | |     | |     |   | | |
A | G A C T G C C C A G G G | G G | G C T G     G A | G A
  | | |           | | | | | | | | |   |   |     |   | | |
  G G A A         | A G G G A G G A   | A G     G   G G A
  | |   |         | | | | | | | |     |   |     |   | | |
  G G C A         C A G G G A G G     C   G     G   G G A
\end{BVerbatim}
}
\caption{Five DNA sequences in a multiple sequence alignment that is rated as `good'. From Figure 3.1 in \cite{wolff-2006}, reproduced with permission.}
\label{ma-dna-figure}
\end{figure}

In this kind of application in biochemistry, two or more DNA sequences are arranged one above another as in the figure (or side by side) and then, by judicious `stretching' of sequences in a computer, the aim is to bring as many symbols as possible into alignment that match each other from one sequence to another.

With most sets of sequences, exhaustive search will not work because there are astronomically many possible alignments. This means that it is necessary to use heuristic methods which will normally deliver solutions that are reasonably good, within a reasonable amount of time.

\subsubsection{An example of an SPMA}\label{spma-example-section}

\begin{figure}[!htbp]
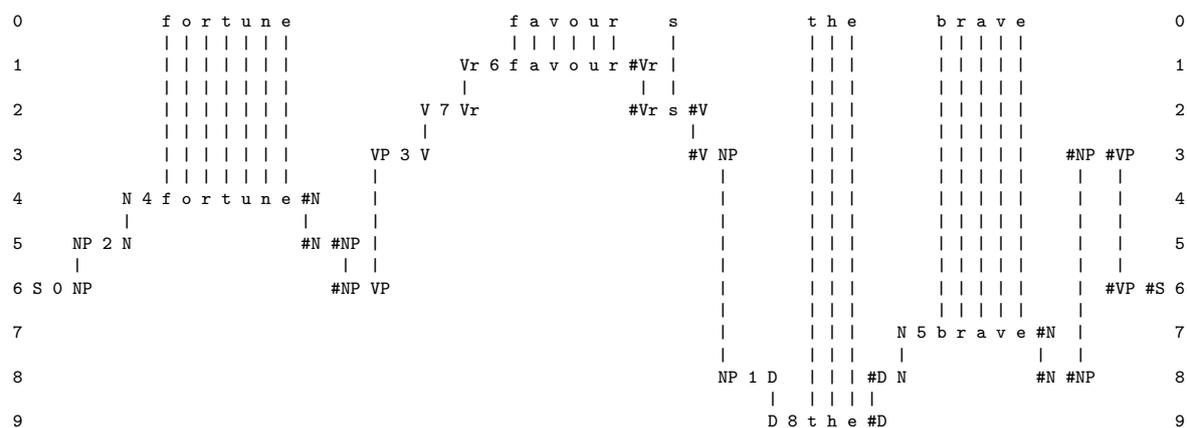

\fontsize{07.00pt}{08.40pt}
\centering
{\bf
\begin{BVerbatim}
0              f o r t u n e                      f a v o u r     s             t h e        b r a v e               0
               | | | | | | |                      | | | | | |     |             | | |        | | | | |
1              | | | | | | |                 Vr 6 f a v o u r #Vr |             | | |        | | | | |               1
               | | | | | | |                 |                 |  |             | | |        | | | | |
2              | | | | | | |             V 7 Vr               #Vr s #V          | | |        | | | | |               2
               | | | | | | |             |                          |           | | |        | | | | |
3              | | | | | | |        VP 3 V                          #V NP       | | |        | | | | |    #NP #VP    3
               | | | | | | |        |                                  |        | | |        | | | | |     |   |
4          N 4 f o r t u n e #N     |                                  |        | | |        | | | | |     |   |     4
           |                 |      |                                  |        | | |        | | | | |     |   |
5     NP 2 N                 #N #NP |                                  |        | | |        | | | | |     |   |     5
      |                          |  |                                  |        | | |        | | | | |     |   |
6 S 0 NP                        #NP VP                                 |        | | |        | | | | |     |  #VP #S 6
                                                                       |        | | |        | | | | |     |
7                                                                      |        | | |    N 5 b r a v e #N  |         7
                                                                       |        | | |    |             |   |
8                                                                      NP 1 D   | | | #D N             #N #NP        8
                                                                            |   | | | |
9                                                                           D 8 t h e #D                             9
\end{BVerbatim}
}
\caption{An SPMA produced by the SP Computer Model with a New SP-pattern (in row 0) representing a sentence to be parsed and a set of Old SP-patterns supplied by the user (including those in rows 1 to 9, one Old SP-pattern per row), each of which represents a grammatical category, and that includes words. Reproduced from Figure 2 in \cite{spneural-2016}, with permission.}
\label{fortune-brave-spma-figure}
\end{figure}

Figure \ref{fortune-brave-spma-figure} shows an example of an SPMA. Here, a New SP-pattern that represents a sentence in natural language (in row 0) is, in effect, analysed into its sections and subsections, in essentially the same manner as parsing in theoretical and computational linguistics, but within the versatile framework of SP-multiple-alignment instead of a hard-wired tree structure.

\newgeometry{left=2cm,right=2cm,top=2cm,footskip=0.1cm}

\restoregeometry

The SP-patterns in rows 1 to 9 of the figure, one SP-pattern per row, are Old SP-patterns, drawn from a much larger repository of Old SP-patterns.

The main features that distinguish an SPMA from a multiple sequence alignment are described in \cite[Section 4]{sp-extended-overview} and \cite[Sections 3.4 to 3.7]{wolff-2006}.

\subsubsection{The creation of SP-multiple-alignments}\label{creation-of-spmas-section}

As with the building of multiple sequence alignments, it is necessary to use heuristic search in the building of SPMAs to obtain reasonably good results in a reasonable time.

The creation of an SPMA like the one shown in Figure \ref{fortune-brave-spma-figure} begins with the New SP-pattern shown in row 0 of that figure and the repository of Old SP-patterns mentioned above which includes the ones shown in rows 1 to 9 of the figure.

At first, each of the Old SP-patterns in the repository is matched with the New SP-pattern as outlined in Section \ref{icmup-section}, below, including the kinds of discontinuous matching outlined in Section \ref{discontinuous-patterns-section}. Each match is evaluated in terms of its potential to compress the New SP-pattern. From the best of those matches, an SPMA is created from the New SP-pattern and one Old SP-pattern.

In subsequent processing, each newly-created SPMA is treated as if it was a single SP-pattern. As such, it may be matched with the New SP-pattern, any of the Old SP-patterns, and any of the other SPMAs in the current run of the program. As before, the best matches are selected and corresponding SPMAs are created, and then the cycle is repeated until no more good matches can be found.

If a good match is found between two `parent' SPMAs, the `child' SPMA that is formed from that match includes all the SP-patterns in both parents. Likewise for a good match between an SPMA and an SP-pattern. In this way, SPMAs with many SP-patterns can build up quickly.

\subsubsection{Versatility of the SP-multiple-alignment construct}\label{versatility of spma section}

The SPMA concept is largely responsible for the versatility of the SP System, which is outlined in Section \ref{strengths-potential-section}, below. In all areas except unsupervised learning (Section \ref{unsupervised-learning-section}), it is almost exclusively responsible for that versatility, but it also has major role in unsupervised learning, together with other processing.

\subsection{Unsupervised learning}\label{unsupervised-learning-section}

In broad terms, unsupervised learning in the SP System means compressing a relatively large body of New SP-patterns from the system's environment to create a smaller body of Old SP-patterns which may be added to the repository of Old SP-patterns, in keeping with the schematic view of the SP System shown in Figure \ref{sp-input-perspective-figure}, Section \ref{sp-abstract-section}. For a given body of New SP-patterns, that smaller body of Old SP-patterns is called its {\em SP grammar}.

It should be mentioned that, although some useful results have been achieved with unsupervised learning in the SP Computer Model (see \cite[Chapter 9]{wolff-2006}), there are some unsolved problems with unsupervised learning in the program, noted in \cite[Section 3.3]{sp-extended-overview}. For those reasons, with the example in Figure \ref{fortune-brave-spma-figure} and examples shown in later sections, it has been necessary to provide the model with appropriate SP-patterns rather than allowing the model to learn those SP-patterns for itself.

As we shall see in Section \ref{the-donsvic-principle-section}, the SP System, via unsupervised learning, can bootstrap a knowledge of granular structures such as words, and grammatical rules from samples of an English-like artificial language in which all punctuation and spaces between words have been removed.

\subsection{Existing and potential strengths of the SP System}\label{strengths-potential-section}

In keeping with the aim of simplifying and integrating observations and concepts across a broad canvass (mentioned at the beginning of Section \ref{sp-outline-section}), the SP System has strengths and potential in several different areas, as summarised in \cite[Section 3.7]{sp-micmup}.

In brief, the strengths and potential of the SP Computer Model in AI include unsupervised learning, pattern recognition, several kinds of reasoning, the processing of natural language, planning, problem solving, and more. Likewise, it has strengths in the representation of several different kinds of knowledge. And because these things all flow largely from the SPMA construct, there is clear potential for the seamless integration of different aspects of AI and different kinds of knowledge, in any combination.

On the strength of this evidence, and evidence summarised in the next two subsections, it seems fair to say that {\em the SP System provides a relatively promising foundation for the development of artificial general intelligence.}

What is said in this chapter about transparency and granularity is likely to apply to the evidence summarised in this section, in the next subsection, and in Section \ref{non-ai-areas-of-application-section}.

\subsubsection{Potential to help solve AI-related problems}label{potential-with-ai-related-problems-section}

Apart from the existing and potential strengths just described, the SP System has clear potential to help solve several problems in AI research. Many of these have been described by leading researchers in AI in interviews with science writer Martin Ford and, after any corrections by the interviewees, reported in Ford's book {\em Architects of Intelligence} \cite{ford-2018}. The potential of the SP System to help solve many of those problems, and some others, is described in \cite{sp-archai-v4}.

\subsubsection{Areas of application apart from AI}\label{non-ai-areas-of-application-section}

\sloppy Apart from AI, the SP System has clear potential in other areas. Relevant papers may be downloaded via links in \href{http://www.cognitionresearch.org/sp.htm}{www.cognitionresearch.org/sp.htm}. They include: the management of big data \cite{sp-big-data}; computer vision and the understanding of natural vision \cite{sp-vision}; the development of intelligent databases \cite{wolff-sp-intelligent-database}; medical diagnosis \cite{wolff-medical-diagnosis}; and more.

\subsection{SP-Neural}\label{sp-neural-section}

The SP System has been developed largely as an abstract model, with well-known features of HLPC as its main touchstones of success. But it is a matter of some interest to discover whether the main features of the SP System may be reproduced with with neural tissue, and if so how.

In the SP programme of research, a first tentative model in this area is called {\em SP-Neural}, described and discussed in \cite{spneural-2016}.

It seems that a case can be made for: modelling SP-symbols with single neurons or, more plausibly, with small clusters of neurons; SP-patterns may be modelled with arrays of neural symbols; and, very tentatively, spike potentials travelling along axons connecting neural SP-symbols and neural SP-patterns may achieve the effect of building SPMAs, and perhaps, unsupervised learning.

As with the non-neural SP System, it seems likely that the creation of a computer model of SP-Neural will help to clarify issues where there is uncertainty at present.

\subsection{Future developments}

\begin{figure}[!htbp]
\centering
\includegraphics[width=0.9\textwidth]{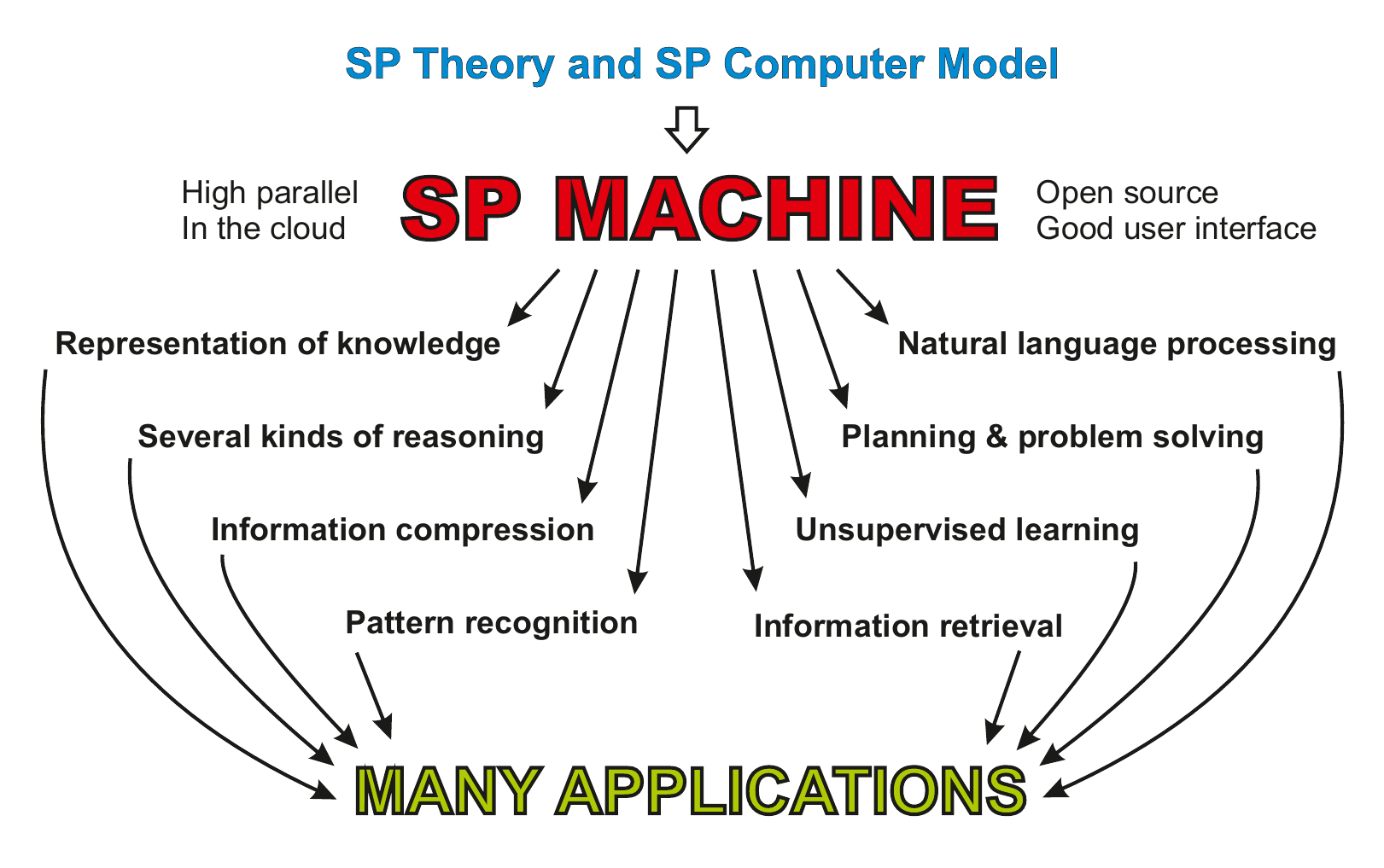}
\caption{Schematic representation of the development and application of the SP Machine. Adapted from Figure 2 in \cite{sp-extended-overview}, with permission.}
\label{sp-machine-figure}
\end{figure}

It is envisaged that the SP Theory of Intelligence and the SP Computer Model will be developed into a highly parallel ``SP-Machine'', as described in \cite{sp-palade-wolff}, and shown schematically in Figure \ref{sp-machine-figure}.

It is envisaged that this will provide a foundation for further work by researchers anywhere, singly or in teams, towards the development of a system with the strength and robustness for large-scale applications.

\section{Information compression and the representation and processing of knowledge in the SP System}\label{ic-kr-pr-in-sp-system-section}

Before getting on to transparency and granularity in the SP System (Sections \ref{transparency-via-audit-trails-section}, \ref{transparency-granularity-familiarity-section}, and \ref{interpretability-explainability-section}), something needs to be said about IC and how it relates to the representation of knowledge in the SP System, and how it is processed.

\subsection{Information compression via the matching and unification of patterns}\label{icmup-section}

A working hypothesis in the SP programme of research is that IC may always be understood as the product of a search for patterns that match each other and the merging or `unifying' of patterns that are the same. The expression ``Information Compression Via the Matching and Unification of Patterns'' will be abbreviated as `ICMUP'.

\begin{figure}[!hbt]
    \centering
    \includegraphics[width=0.9\textwidth]{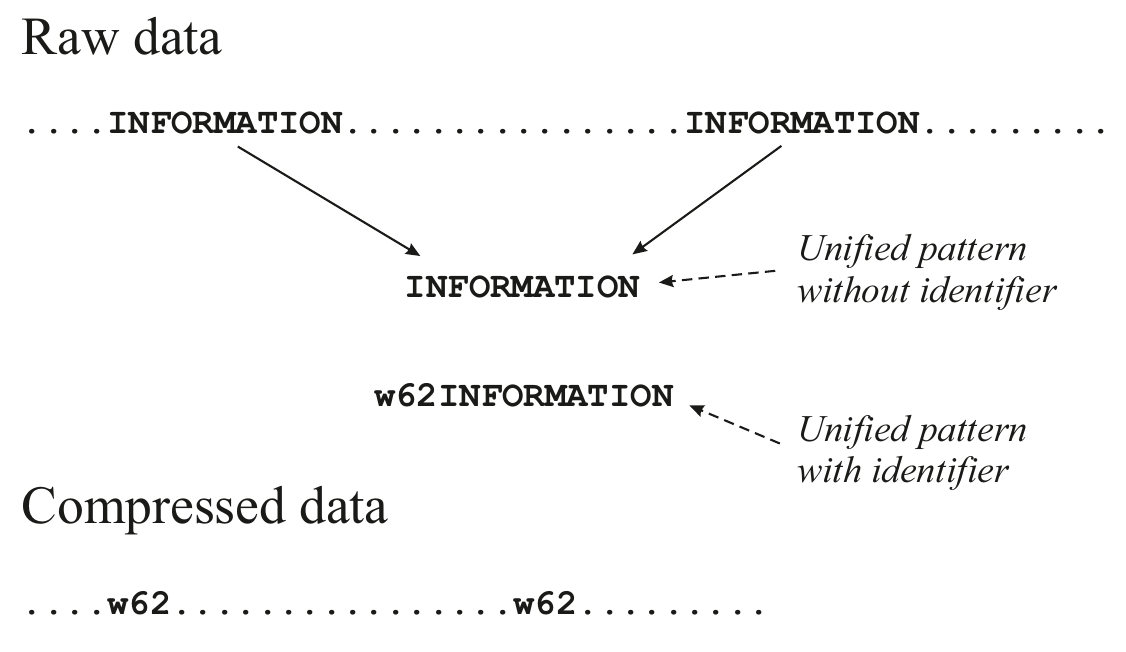}
    \caption{How two instances of the pattern `\texttt{INFORMATION}' in a body of raw data may be merged to form a single `unified' pattern or `chunk' of information, shown below the `raw data'. The rest of the figure is considered later. Adapted from Figure 2.3 in \protect\cite{wolff-2006}, with permission.}
    \label{unification-figure}
\end{figure}

This idea is illustrated in the upper part of Figure \ref{unification-figure}. Here, in some `raw data' shown at the top of the figure, two examples of the pattern `\texttt{INFORMATION}' are merged or unified to create a single instance, shown immediately below.

In this instance, there is compression of information because two instances of `\texttt{INFORMATION}' are reduced to one. The rest of the figure is considered in Section \ref{chunking-with-codes-icmup-section}, below.

To achieve IC via ICMUP in a given body of raw information, {\bf I}, the patterns that are to be unified must be relatively frequent in {\bf I} and no less frequent than would occur by chance.

That figure for chance varies inversely with the size of the pattern, so that, with a given {\bf I}, a large pattern like this:

\begin{quote}

    `\texttt{pneumonoultramicroscopicsilicovolcanoconiosis}'

\end{quote}

\noindent may exceed the threshold with a frequency as low as 2, while smaller patterns like `\texttt{si}' may only reach the threshold with a higher frequency.

\subsection{Discontinuous patterns}\label{discontinuous-patterns-section}

An important point to mention in connection with ICMUP is that the concept of `pattern' in the SP programme of research includes patterns that are `discontinuous' in the sense that they may be interwoven with other information.

For example, a pattern like `\texttt{ABC}' may be seen in `\texttt{LMANOPBQCSTU}', and likewise in many other sequences.

From prominent features of HLPC, such as the way we can recognise a familiar pattern like a car despite interfering patterns such as iron railings or the branches of a plant, it has been understood from the beginning of the SP research that there would be a need for that kind of capability in the SP Computer Model.

A first goal was to create a system that could recognise good full and partial matches between sequences (where `good' means good in terms of IC), and could deliver two or more such solutions where they exist. The method which has been adopted, and incorporated in successive versions of the SP Computer Model, is described in \cite[Appendix A]{wolff-2006}.

\subsection{Seven variants of ICMUP}\label{variants-of-icmup-section}

ICMUP is an intrinsically simple idea, but it comes in seven main variants which add a lot in terms of its descriptive and explanatory value. The variants are described in \cite[Section 5]{sp-micmup} and \cite[Sections 6, 7, and 8]{sp-compression}, and are outlined more briefly here.

In general, these structures would arise from IC via unsupervised learning, as outlined in Section \ref{unsupervised-learning-section}.

In general, these structures that are very widely used and are likely to be familiar to most people. Some comments to that effect are in Section \ref{transparency-granularity-familiarity-section}.

\subsubsection{Basic ICMUP}\label{basic-icmup-section}

{\em Basic ICMUP} is our first variant of ICMUP, essentially what is described at the beginning of Section \ref{icmup-section}: if two or more patterns match each other, they may be unified to create one copy, with a corresponding compression of information.

\subsubsection{Chunking-with-codes and ICMUP}\label{chunking-with-codes-icmup-section}

A problem with Basic ICMUP is that, for each set of unified patterns, information is lost about their {\em locations} in {\bf I}, except for the unified pattern itself, assuming it retains a place in {\bf I}.

A solution to this problem, called {\em Chunking-with-codes}, is to assign a relatively short identifier or {\em code} to the unified pattern---which is commonly referred to as a {\em chunk} of information---and then to place a copy of the chunk, with its code, in a separate `dictionary' of patterns. Then replace each copy of the chunk in {\bf I} with the code for the chunk. This is illustrated in the lower part of Figure \ref{unification-figure}, where the chunk of information is `\texttt{INFORMATION}' and the code for that chunk is `\texttt{w62}'.

Because in general the code should be smaller than the chunk it represents, there should be an overall compression of {\bf I}.

\subsubsection{Schema-plus-correction and ICMUP}\label{schema-plus-correction-icmup-section}

An interesting variant of MICMUP is known as {\em Schema-plus-correction}. Perhaps the best-known example is a menu of dishes that are available in a cafe or restaurant.

The menu itself may be seen as special kind of `chunk' of information which, as with chunking-with-codes, has a relatively short name, identifier, or `code'---something like `Menu', `Your choices', `As you like it', and so on.

What makes it different from an ordinary example of chunking-with-codes is that it provides the means of introducing variations into the chunk.

A typical menu offers three or more places where variations may be introduced. These would be parts of the menu such as `starter', `main course', and `pudding'. With each of these there would be a selection of dishes that the diner may choose, such as `soup', `antipasto', and so on for the starter, `vegan chickpea curry', `shepherd's pie', and so on for the main course, and `ice cream', `apple crumble', and so on for the pudding.

\subsubsection{Run-length encoding and ICMUP}\label{run-length-coding-icmup-section}

Run-length encoding may be applied where there is a sequence of two or more matching patterns, each one contiguous with next one. Then a sequence like `\texttt{INFORMATIONINFORMATIONINFORMATIONINFORMATIONINFORMATION}' may be reduced to something like `\texttt{INFORMATION} $\times 5$', or, more generally as `\texttt{INFORMATION}*', where the `*' indicates that the given pattern is repeated but without specifying how many times it is repeated.

\subsubsection{Part-whole hierarchies and ICMUP}\label{part-whole-hierarchies-icmup-section}

A fifth variant of ICMUP is the way things can be organised as part-whole hierarchies. A car may be seen to be divided to engine, wheels, body, and so on. And each of these things may be divided into parts and subparts, and so on.

Economies arise because, at any one level in a part-whole hierarchy, all the alternatives at that level share the same place in the hierarchy, which saves having to repeat that information for every one of the alternatives.

For example, someone buying a particular model of a car may be offered a choice of two or three different engines. Each of the alternatives may be described without the need, for each alternative, to describe the rest of the car. Hence, those several copies of the context of `engine' have been merged into a single copy, in accordance with ICMUP.

\subsubsection{Class-inclusion hierarchies and ICMUP}\label{class-inclusion-hierarchies-section}

One of the meanings of the word `class' is that it is a collection of things that share certain features. So the class `table' applies to things that have a horizontal top that may be used as a temporary place to put things, especially plates, knives and forks and so on at meal times, often with four legs, often made of wood, and so on.

This may be seen as an example of ICMUP because, across all the many examples of tables, the features that they have in common have been seen to match each other and have been unified to create the list of features for the class `table'.

It is true that there are many exceptions and special cases---for example, not all tables are made of wood---but that does not alter the great economies that can be achieved, in both thinking and communication, from the use of classes like `table'. The class saves having to describe all the features of a table every time one wants to talk about tables or simply remember something about tables, such as the way a table may be used to help in the changing of a light bulb.

From this idea of a class, it is a short step to the idea of a hierarchy of subclasses, subsubclasses, and so on. At each level in the hierarchy, there are features that are inherited by all the higher levels.

\subsubsection{SP-multiple-alignment and ICMUP}\label{spma-icmup-section}

The last of the seven variants of ICMUP is the concept of SPMA that has been described already in Section \ref{concept-of-spma-section}.

Out of the seven variants of ICMUP, it appears that SPMA can be, with appropriate data, the most effective means of compressing information, largely because the matching and unification of patterns may occur at several different levels, not just one level.

And this seventh variant of ICMUP has a special status amongst the seven variants because SPMA may be seen to encompass all of the other six variants, and, within any one SPMA, there can be a seamless integration of the other six variants.

It appears that it is this versatility which is largely responsible for the versatility of SPMAs in modelling diverse aspects of intelligence, in the representation of diverse kinds of knowledge, and in the seamless integration of aspects of intelligence and kinds of knowledge, in any combination (Section \ref{strengths-potential-section}).

\subsection{The DONSVIC principle}\label{the-donsvic-principle-section}

An idea which is fundamental in the workings of the SP System is the `DONSVIC' principle, meaning the ``Discovery Of Natural Structures Via Information Compression'' It is described quite fully in \cite[Section 5.2]{sp-extended-overview}.

It seems that the reason that IC does not normally have the effect of revealing `natural' structures is that, largely because of the low power of early computers, most systems for IC have been designed to be `quick and dirty', sacrificing accuracy for speed on those low-powered computers. Now that computers are more powerful, one can be more ambitious.

The same section of the paper \cite[Section 5.2]{sp-extended-overview} describes how the MK10 program for unsupervised learning of segmental structures---with IC as its driving principle---may discover structures in natural language such as words and phrases, and this without any prior knowledge of any of those structures, and without any markers in the raw data such as punctuation and spaces between words to show the beginnings and ends of segmental structures \cite{wolff-1988}. And in a similar way, the SNPR program for unsupervised learning of grammars---with IC as its central principle---demonstrates successful learning of the grammars of English-like artificial languages ({\em ibid}).

In that connection, there is evidence that a first language can be learned by children without `reinforcement' as normally understood, or any other kind of explicit teaching or the correction of errors (see, for example, \cite{lenneberg-1962,brown-2014,chater-vitanyi-2007}). It seems likely that the same applies to the learning of non-syntactic structures as well.

As a rough generalisation, structures that may be discovered via the DONSVIC principle from a given body of information, {\bf I}, are ones that are useful in compressing {\bf I} and are likely to be useful in compressing any later body of information that has a similar structure.

These observations are in accord with substantial evidence for the significance of IC in HLPC \cite{sp-compression}.

\subsubsection{The DONSVIC principle and granularity}\label{donsvic-granularity-section}

It is assumed in this research that, in HLPC, the DONSVIC principle applies to the unsupervised learning of the great majority of entities, structures, or events, that we recognise `naturally', including the kinds of SP-patterns shown in rows 1 to 9 of Figure \ref{fortune-brave-spma-figure}.

If it is accepted that most of the ``entities, structures, or events, that we recognise `naturally'\thinspace'' would also qualify as `granules', we should also accept that granules and the ways in which they are structured are likely to emerge via learning processes that conform to the DONSVIC principle, either in human brains, or in artificial unsupervised learning of the future.

\subsubsection{The DONSVIC principle and familiarity}\label{donsvic-familiarity-section}

In the same way that the DONSVIC principle suggests that information granules and their structures emerge from a search for patterns with an optimum combination of size and frequency, it is likely that the way in which those granules are structured (as described in Section \ref{variants-of-icmup-section}) will also be familiar to people.

The familiarity of those kinds of structures---such as chunking-with-codes, run-length coding, part-whole hierarchies, and class-inclusion hierarchies---will clearly be important in ensuring the interpretability, explainability, and transparency of knowledge created via unsupervised learning in the SP System, and via the building of SPMAs.

\subsection{Ideas related to the concept of a granule}\label{related-ideas-section}

This section briefly discusses two ideas that appear to be relevant to the concept of a granule, and also to key ideas in the SP System.

\subsubsection{The concept of a {\em chunk} of information}\label{miller-chunk-section}

As we have seen in Sections \ref{chunking-with-codes-icmup-section} and \ref{schema-plus-correction-icmup-section}, and elsewhere above, the concept of `chunk' can be useful in describing any small coherent body of information. As such, it is similar to the concept of a `granule'.

It appears that the concept of a `chunk' in cognition-related research, was first introduced in George Miller's much-quoted paper on ``The magical number seven, plus or minus two'' \cite{miller-1956} where he argued that:

\begin{quote}

    ``...~we must recognize the importance of grouping or organizing [information] into units or chunks. Since the memory span [of a typical person] is a fixed number of chunks, we can increase the number of bits of information that it contains simply by building larger and larger chunks, each chunk containing more information than before.'' \cite[p.~93]{miller-1956}.

\end{quote}

In keeping with that description, Section \ref{icmup-section}, above, suggests how chunks of information may be discovered via the matching and unification of patterns.

Since Miller's seminal paper, the concept of a {\em chunk} of information has been and still is widely used in the academic literature in cognitive science and cognitive psychology. Now the word `chunk', apparently without reference to Miller's concept, has been associated with the word `granule' like this:

\begin{quote}

    ``Granular computing ...~is a research area focused on representing, reasoning, and processing basic {\em chunks} of information, namely {\em granules}.'' \cite[p.~1835]{fujita-etal-2018}, emphasis added.

\end{quote}

Apart from that connection between `chunk' and `granule', a search of the literature suggests that there is at present little interest in examining possible synergies between the two areas of research.

As noted in Section \ref{sp-patterns-and-symbols-section}, the concept of an `SP-pattern' in the SP System appears to capture much of what is meant by the concepts of `granule' and `chunk'.

\subsubsection{Object-oriented programming}\label{oo-section}

Another thing with much of the flavour of the concepts of `granule', `SP-pattern', and `chunk', is the concept of a discrete entity or object in object-oriented programming. From small beginnings in Norway \cite{birtwistle-etal-1973}, this paradigm for programming has grown to be a widely-adopted feature in the design of programming languages, and in the design of software systems.

For readers not already familiar with OO-programming and OO-design, the neat idea is that a software system should be a model of the system it is to serve, with a discrete software `object' for each entity or object in the system to be modelled, and with hierarchies of `classes' of object and with `inheritance' of `attributes' of objects from higher levels to lower levels (Section \ref{class-inclusion-hierarchies-section}).

Some connections have been made between concepts of granularity and object-oriented programming (eg \cite{lee-liou-1996}) but it appears not to be a live issue.

\subsection{Tying things together?}\label{tying-things-together-section}

In view of what has been said earlier about information granules (Section \ref{granularity-introduction-section}), about SP-patterns (Section \ref{sp-patterns-and-symbols-section}), about information chunks (Section \ref{miller-chunk-section}), and about entities or objects in object-oriented programming (Section \ref{oo-section}), there seems to be a case for tying these concepts together, perhaps within the framework of information compression. It seems likely that the SP System could accommodate them all.

\section{Transparency via audit trails}\label{transparency-via-audit-trails-section}

Compared with DNNs, the SP System has the striking advantage that it provides a full audit of what it is doing, and it provides it in a form that people can understand. That advantage applies to unsupervised learning in the SP System, which means the creation of SP-grammars (Section \ref{unsupervised-learning-section}), and it also applies to the building of SPMAs by the SP System (Section \ref{creation-of-spmas-section}), which provides the means of modelling all the other AI-related things that the SP System can do, such as the processing of natural language, recognition of entities, several forms of reasoning, and so on, as summarised in Section \ref{strengths-potential-section}.

\begin{figure}[!htbp]
\centering
\includegraphics[width=0.8\textwidth]{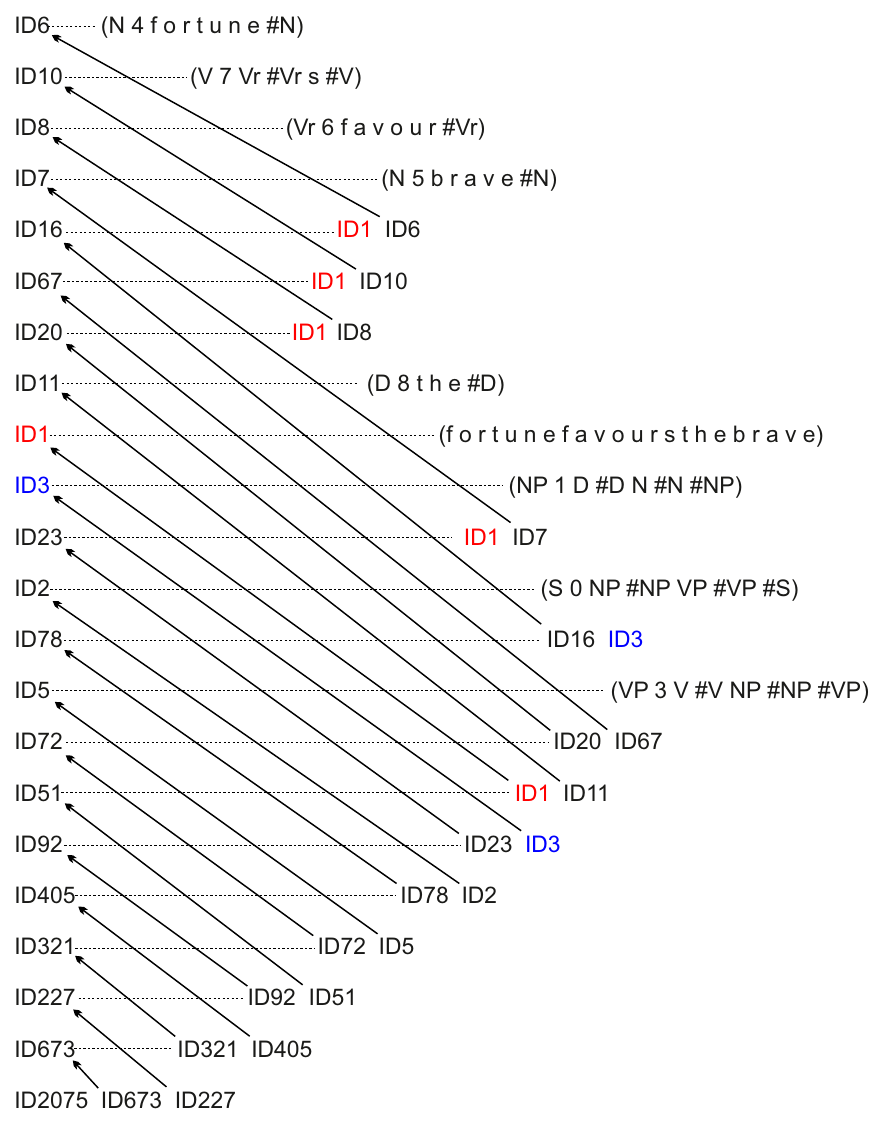}
\caption{An audit trail for the creation of the SPMA shown in Figure \ref{fortune-brave-spma-figure}. It should be interpreted as follows: the figure should be read from the bottom to the top, starting with a row about the SPMA in Figure \ref{fortune-brave-spma-figure}; at the beginning of each row there is an identifier for an SPMA or an SP-pattern; in each row that describes an SPMA, there are two identifiers to the right, referencing the two structures that were matched and unified to create the given SPMA; those two structures might be one SP-pattern with another SP-pattern (or different parts of itself), or an SPMA with an SP-pattern (in either order), or an SPMA with an SPMA; in most cases, there is an arrow from each of the two identifiers to the SPMA or SP-pattern that it refers to; but with the SP-patterns `ID1' and `ID3', each of their identifiers (shown in colour) appears more than once in the figure, so to avoid undue clutter in the figure in each of those two cases, only one arrow is shown.}
\label{audit-trail-figure}
\end{figure}

Figure \ref{audit-trail-figure} shows a bare-bones audit trail for the creation of the SPMA shown in Figure \ref{fortune-brave-spma-figure}, to illustrate how, for each SPMA that is created by the SP Computer Model, the program provides detailed information about how the SPMA is built. The caption to Figure \ref{audit-trail-figure} describes how it should be interpreted.

\newgeometry{left=2cm,right=2cm,top=2cm,footskip=0.1cm}

\restoregeometry

The information in Figure \ref{audit-trail-figure} is only an extract from the much fuller information that the SP Computer Model provides. With each SPMA referenced in the figure, and many others on paths that turned out to be blind alleys, the following information is provided:

\begin{itemize}

    \item The full structure of the SPMA.

    \item The pairing that produced the given SPMA: an SP-pattern with an SP-pattern, or an SPMA with an SP-pattern (in either order), or an SPMA with an SPMA (see Section \ref{creation-of-spmas-section}).

    \item A full evaluation of each SPMA in terms of IC, described in detail in \cite[Section 3.5]{wolff-2006} and \cite[Section 4.1]{sp-extended-overview}.

    \item Absolute and relative probabilities associated with each SPMA, calculated as shown in \cite[Section 3.7]{wolff-2006} and \cite[Section 4.4]{sp-extended-overview}.

\end{itemize}

A similar level of detail is provided for the creation of SP-grammars by the SP Computer Model.

As noted above, this kind of transparency in the workings of the SP Computer Model contrasts with the considerable obscurity in the workings of DNNs (Section \ref{transparency-introduction-section}).

\section{Transparency via granularity and familiarity}\label{transparency-granularity-familiarity-section}

As we have seen in Section \ref{the-donsvic-principle-section}, it appears that the concepts of granularity and the DONSVIC principle are closely related. For that reason, they will be discussed together in subsections below which describe how the SP System exhibits granularity associated with each of the seven variants of ICMUP described in Section \ref{variants-of-icmup-section}.

In addition to granularity, it seems that, because they are so widely used, each of those variants of ICMUP are likely to strike a chord of familiarity with most people.

Thus because of both granularity and familiarity, each of those seven variants of ICMUP are likely to contribute to interpretability and explainability, and a consequent transparency, in the operation of the SP System.

\subsection{Granularity, familiarity, and Basic ICMUP}\label{granularity-basic-icmup-section}

Although Basic ICMUP, as described in Section \ref{basic-icmup-section}, is indeed remarkably basic and simple, it is also surprisingly powerful, providing examples of both granularity and the DONSVIC principle. It may, for example, be seen to provide the main mechanism for our perception of the world as being populated by three-dimensional objects.

In case this seems obscure, the learning and perception of 3D objects is a development envisioned for the SP System, described in \cite[Sections 6.1 and 6.2]{sp-vision}, but it has not yet been implemented in the SP Computer Model.

In brief, the basic idea is that, with any kind of object that is new to us, we may view it from several different angles so that there is overlap between neighbouring fields of view. Then our brains can piece together a three-dimensional model of the object by merging the overlapping areas (via Basic ICMUP), much as a panoramic view of a scene may be made with a digital camera from several overlapping views of the scene. In a similar way, we may recognise 3D objects that we already know, and refine our knowledge of them.

The idea of creating 3D models from two or more views, called `photogrammetry', is the basis of commercial and free systems that are available for creating 3D models from photographs.\footnote{See, for example, Agisoft (\href{https://www.agisoft.com/}{www.agisoft.com/}), All3DP (\href{https://all3dp.com/}{all3dp.com/}), Sculpteo (\href{https://www.sculpteo.com}{www.sculpteo.com}), and more.}

The same kinds of process are at work in the creation of Google's `Streetview'. Here, overlapping digital photographs taken of streets and junctions all over the world are pieced together to create very useful 3D digital models of those many streets and junctions.

Because objects are such a familiar aspect of how we perceive the world, Basic ICMUP contributes to familiarity as well as granularity, and both of them contribute to transparency in results produced by the SP System.

Even without objects, A little reflection will show that Basic ICMUP is something we do all the time. Whenever we recognise something that we know already, we are employing Basic ICMUP. And whenever we see something new but see a similarity to something we know already, we are employing Basic ICMUP.

\subsection{Granularity, familiarity, and chunking-with-codes}\label{granularity-chunking-with-codes-section}

From the perspective of the chunking-with-codes principle for achieving IC (Section \ref{chunking-with-codes-icmup-section}), it seems that a `chunk' of information qualifies as an information granule, and as an example of the DONSVIC principle at work.

It appears that chunking-with-codes is widespread in HLPC, especially in our use of language. For example, a word like `house' may be understood as a relatively short identifier or code for the much larger chunk of information which is the meaning of `house'. It matters not that the chunk may be relatively generalised so that it accommodates many of the different kinds of houses that people live in. Regardless of the complexity of that concept, the word `house' serves as a short identifier for the concept.

A little reflection shows that, in any natural language, every noun, adjective, verb, and adverb, is in effect a short code for the much larger chunk of information which is what the word means.

Because this method for the economical encoding of information is so simple and effective, it is likely that non-verbal aspects of our thinking would be encoded in a similar way---although it is more difficult to obtain direct evidence for this than it is with the surface forms of natural languages.

As with Basic MUP, chunking-with-codes is an extremely common feature of how we conceive the world, and thus it exhibits familiarity as well as granularity, and both of them contribute to transparency.

\subsection{Granularity, familiarity, and schema-plus-correction}\label{granularity-schema-plus-correction-section}

With regard to schema-plus-correction as a means of achieving IC (Section \ref{schema-plus-correction-icmup-section}), granularity and the DONSVIC principle may be seen at work at two main levels: the schema itself and the `chunks' of information which serve as `corrections' to the schema.

As we saw in Section \ref{schema-plus-correction-icmup-section}, a menu in a restaurant or cafe is a good example of the schema-plus-correction means of achieving IC. This is illustrated in Figure \ref{menu-grammar-figure}, which is a relatively simple example of an SP-grammar composed of SP-patterns. The first SP-pattern, `\texttt{MU ST \#ST MC \#MC PD \#PD \#MU}', represents an outline of the menu with slots for the starter (`\texttt{ST \#ST}'), main course (`\texttt{MC \#MC}'), and pudding (`\texttt{PD \#PD}'), and all the other SP-patterns represent different dishes in those three categories.

\begin{figure}[!htbp]
\fontsize{09.00pt}{10.80pt}
\centering
{\bf
\begin{BVerbatim}
MU ST #ST MC #MC PD #PD #MU  | Prepare meal
ST 0 mussels #ST             | Starter: mussels
ST 1 soup #ST                | Starter: soup
ST 2 avocado #ST             | Starter: avocado
MC 0 lasagna #MC             | Main course: lasagna
MC 1 beef #MC                | Main course: beef
MC 2 nut-roast #MC           | Main course: nut-roast
MC 3 kipper #MC              | Main course: kipper
MC 4 salad #MC               | Main course: salad
PD 0 ice cream #PD           | Pudding: ice cream
PD 1 apple-crumble #PD       | Pudding: apple crumble
PD 2 fresh-fruit #PD         | Pudding: fresh fruit
PD 3 tiramisu #PD            | Pudding: tiramisu
\end{BVerbatim}
}
\caption{An SP grammar composed of SP-patterns that represent a three-course meal. Each SP-pattern has a comment to the right which explains what it is about, with the marker `\texttt{|}' at the beginning of each comment. Adapted from Figure 2 in \cite{sp-software-engineering}.}
\label{menu-grammar-figure}
\end{figure}

How this grammar may be used in practice may be seen in the SPMA shown in Figure \ref{menu-spma-figure}. A prominent difference between this SPMA and the one shown earlier in Figure \ref{fortune-brave-spma-figure} is that, in the earlier one, SP-patterns are arranged horizontally, while in the later one, they are arranged vertically. This has no theoretical significance and is purely a matter of which arrangement is the best fit for the page.

\begin{figure}[!htbp]
\fontsize{09.00pt}{10.80pt}
\centering
{\bf
\begin{BVerbatim}
0     1     2               3         4

MU -- MU                                           | Menu
      ST ------------------ ST                     | Starter
0 ------------------------- 0
                            mussels
      #ST ----------------- #ST
      MC ---------------------------- MC           | Main course
4 ----------------------------------- 4
                                      salad
      #MC --------------------------- #MC
      PD -- PD                                     | Pudding
1 --------- 1
            apple-crumble
      #PD - #PD
#MU - #MU

0     1     2               3         4
\end{BVerbatim}
}
\caption{The best SPMA created by the SP Computer Model with the New SP-pattern, `\texttt{MU 0 4 1 \#MU}', and the set of Old SP-patterns shown in Figure \ref{menu-grammar-figure}. Adapted from Figure 3 in \cite{sp-software-engineering}.}
\label{menu-spma-figure}
\end{figure}

The SPMA shown in Figure \ref{menu-spma-figure}, is the best of several alternatives created by the SP Computer Model, starting with the New SP-pattern `\texttt{MU 0 4 1 \#MU}' (shown in column 0), and the SP-grammar shown in Figure \ref{menu-grammar-figure}. Old SP-patterns selected from that grammar appear in columns to to 4 in the figure, one SP-pattern per column.

The New SP-pattern, `\texttt{MU 0 4 1 \#MU}', is an economical description of a meal: the SP-symbols `\texttt{MU}' and `\texttt{\#MU}' at the top and bottom show that the New SP-pattern is about the menu, the full version of which is shown in the Old SP-pattern in Column 1; the SP-symbol `\texttt{0}' in column 0 shows that the starter is a dish of mussels; the SP-symbol `\texttt{4}' shows that the main course is a salad; and the SP-symbol `\texttt{1}' shows that the pudding is apple crumble.

Things like menus in cafes and restaurants are not quite as familiar as chunking-with-codes but they are very much part of everyday life and so that we may say that they contribute to familiarity as well as granularity, both them promoting transparency in the workings of the SP System.

\subsection{Granularity, familiarity, and run-length encoding}\label{granularity-run-length-coding-section}

With the repeated instances of `\texttt{INFORMATION}' in Section \ref{run-length-coding-icmup-section}, illustrating the run-length coding concept, repetition of the pattern `\texttt{INFORMATION}' would in itself suggest that it conforms to the DONSVIC principle, and would thus qualify as an information granule.

Whenever a sports coach, for example, says ``keep doing push-ups until I say stop'', he or she is employing run-length coding. It is very widely used and may thus contribute to familiarity and transparency for users of an SP System.

\subsection{Granularity, familiarity, and part-whole hierarchies}\label{granularity-part-whole-section}

\begin{figure}[!htbp]
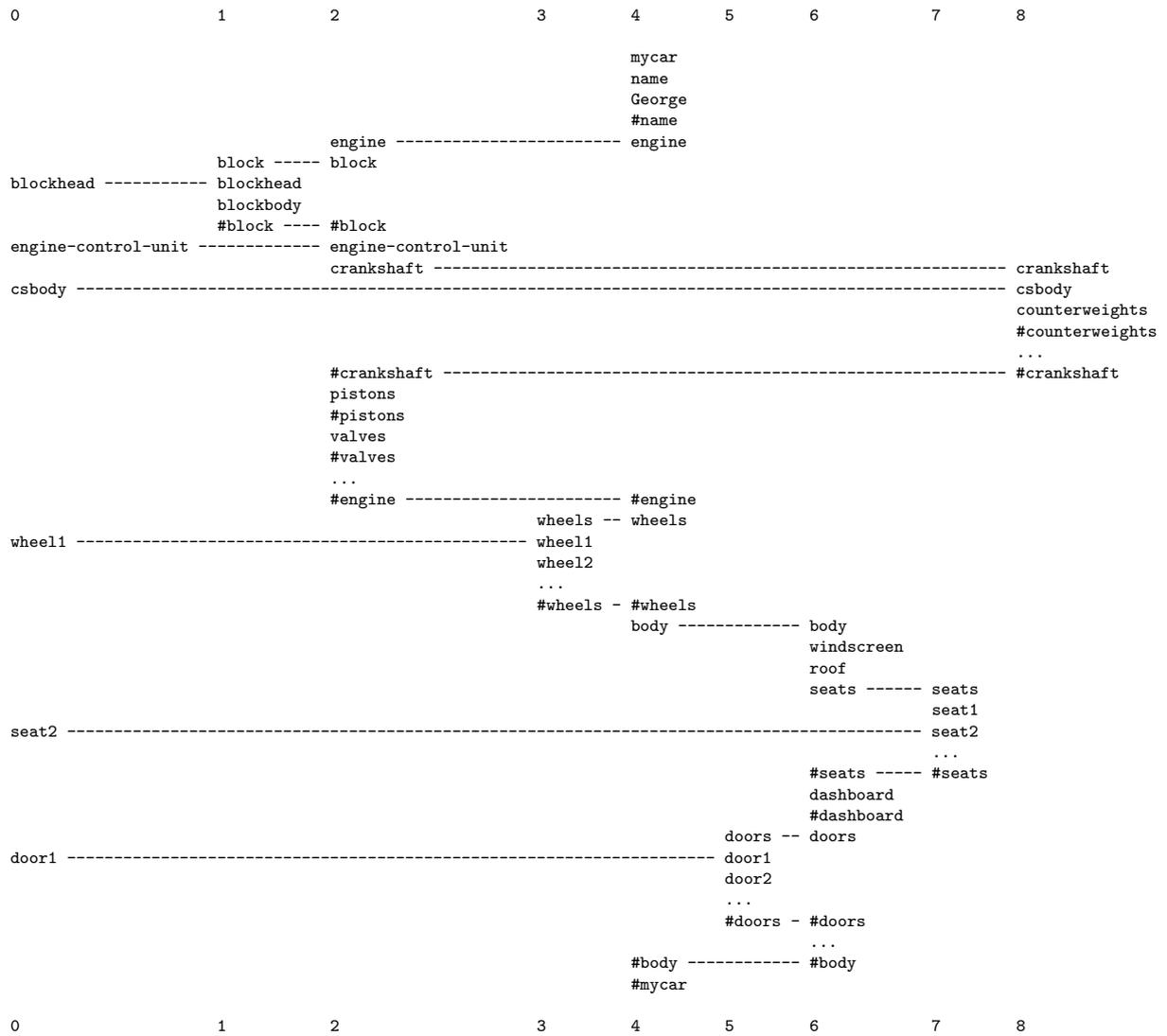

\fontsize{07.00pt}{08.40pt}
\centering
{\bf
\begin{BVerbatim}
0                     1           2                     3         4         5        6            7        8

                                                                  mycar
                                                                  name
                                                                  George
                                                                  #name
                                  engine ------------------------ engine
                      block ----- block
blockhead ----------- blockhead
                      blockbody
                      #block ---- #block
engine-control-unit ------------- engine-control-unit
                                  crankshaft ------------------------------------------------------------- crankshaft
csbody --------------------------------------------------------------------------------------------------- csbody
                                                                                                           counterweights
                                                                                                           #counterweights
                                                                                                           ...
                                  #crankshaft ------------------------------------------------------------ #crankshaft
                                  pistons
                                  #pistons
                                  valves
                                  #valves
                                  ...
                                  #engine ----------------------- #engine
                                                        wheels -- wheels
wheel1 ------------------------------------------------ wheel1
                                                        wheel2
                                                        ...
                                                        #wheels - #wheels
                                                                  body ------------- body
                                                                                     windscreen
                                                                                     roof
                                                                                     seats ------ seats
                                                                                                  seat1
seat2 ------------------------------------------------------------------------------------------- seat2
                                                                                                  ...
                                                                                     #seats ----- #seats
                                                                                     dashboard
                                                                                     #dashboard
                                                                            doors -- doors
door1 --------------------------------------------------------------------- door1
                                                                            door2
                                                                            ...
                                                                            #doors - #doors
                                                                                     ...
                                                                  #body ------------ #body
                                                                  #mycar

0                     1           2                     3         4         5        6            7        8
\end{BVerbatim}
}
\caption{The SPMA shown here is the best of several that the SP Computer Model has created, beginning with several one-SP-symbol SP-patterns (in column 0) that describe some features of a car, and a grammar of Old patterns, some of which are the SP-patterns shown in columns 1 to 8. These SP-patterns include one for `\texttt{mycar}', in column 4, and other SP-patterns that describe parts and sub-parts of `\texttt{mycar}'.}
\label{part-whole-figure}
\end{figure}

A part-whole hierarchy is similar in some respects to the schema-plus-correction example shown in Figures \ref{menu-grammar-figure} and \ref{menu-spma-figure}. Perhaps the main difference is that a part-whole hierarchy would normally have more levels, as in the SPMA in Figure \ref{part-whole-figure}.

\newgeometry{left=2cm,right=2cm,top=2cm,footskip=0.1cm}

\restoregeometry

With many simplifications, this SPMA shows how the SP Computer Model may create an analysis of `\texttt{mycar}' into a part-whole hierarchy when it is presented with some features of `\texttt{mycar}' in column 0 and a repository of Old SP-patterns which include those SP-patterns in columns 1 to 8 of Figure \ref{part-whole-figure}.

As with other examples in this chapter, it is clear that there is granularity in the SP-patterns shown in the figure because economies can be achieved as described in Section \ref{part-whole-hierarchies-icmup-section}, and thus the SPMA is likely to conform to the DONSVIC principle as described in Section \ref{the-donsvic-principle-section}.

As with other variants of ICMUP, part-whole hierarchies are very familiar in everyday life, and that familiarity is likely to contribute to transparency in results from the SP System.

\subsection{Granularity, familiarity, and class-inclusion hierarchies}\label{granularity-class-inclusion-section}

Figure \ref{class-part-plant-figure} shows an SPMA created by the SP Computer Model which, via classes and subclasses of plants, illustrates the concept of a class-inclusion hierarchy, as described in Section \ref{class-inclusion-hierarchies-section}.

Although the categories used by botanists to classify plants have a formal status, it is likely that they have a foundation in what seems `natural', which is itself one facet of the DONSVIC principle (Section \ref{the-donsvic-principle-section}). More generally, categories like that may be seen as information granules.

\newgeometry{left=2cm,right=2cm,top=2cm,footskip=0.1cm}

\begin{figure}[!htbp]
\fontsize{06.00pt}{07.20pt}
\centering
{\bf
\begin{BVerbatim}
0                 1                2                  3              4              5                  6

                  <species>
                  acris
                  <genus> ---------------------------------------------------------------------------- <genus>
                  Ranunculus ------------------------------------------------------------------------- Ranunculus
                                                                                    <family> --------- <family>
                                                                                    Ranunculaceae ---- Ranunculaceae
                                                                     <order> ------ <order>
                                                                     Ranunculales - Ranunculales
                                                      <class> ------ <class>
                                                      Angiospermae - Angiospermae
                                   <phylum> --------- <phylum>
                                   Plants ----------- Plants
                                   <feeding>
has-chlorophyll ------------------ has-chlorophyll
                                   photosynthesises
                                   <feeding>
                                   <structure> ------ <structure>
                                                      <shoot>
<stem> ---------- <stem> ---------------------------- <stem>
hairy ----------- hairy
</stem> --------- </stem> --------------------------- </stem>
                  <leaves> -------------------------- <leaves>
                  compound
                  palmately-cut
                  </leaves> ------------------------- </leaves>
                                                      <flowers> ------------------- <flowers>
                                                                                    <arrangement>
                                                                                    regular
                                                                                    all-parts-free
                                                                                    </arrangement>
                  <sepals> -------------------------------------------------------- <sepals>
                  not-reflexed
                  </sepals> ------------------------------------------------------- </sepals>
<petals> -------- <petals> -------------------------------------------------------- <petals> --------- <petals>
                                                                                    <number> --------- <number>
                                                                                                       five
                                                                                    </number> -------- </number>
                  <colour> -------------------------------------------------------- <colour>
yellow ---------- yellow
                  </colour> ------------------------------------------------------- </colour>
</petals> ------- </petals> ------------------------------------------------------- </petals> -------- </petals>
                                                                                    <hermaphrodite>
<stamens> ------------------------------------------------------------------------- <stamens>
numerous -------------------------------------------------------------------------- numerous
</stamens> ------------------------------------------------------------------------ </stamens>
                                                                                    <pistil>
                                                                                    ovary
                                                                                    style
                                                                                    stigma
                                                                                    </pistil>
                                                                                    </hermaphrodite>
                                                      </flowers> ------------------ </flowers>
                                                      </shoot>
                                                      <root>
                                                      </root>
                                   </structure> ----- </structure>
<habitat> ------- <habitat> ------ <habitat>
meadows --------- meadows
</habitat> ------ </habitat> ----- </habitat>
                  <common-name> -- <common-name>
                  Meadow
                  Buttercup
                  </common-name> - </common-name>
                                   <food-value> ----------------------------------- <food-value>
                                                                                    poisonous
                                   </food-value> ---------------------------------- </food-value>
                                   </phylum> -------- </phylum>
                                                      </class> ----- </class>
                                                                     </order> ----- </order>
                                                                                    </family> -------- </family>
                  </genus> --------------------------------------------------------------------------- </genus>
                  </species>

0                 1                2                  3              4              5                  6
\end{BVerbatim}
}
\caption{An SP-multiple-alignment created by the SP Computer Model. It is the best of several alternatives that the program creates, starting with a set of of New SP-patterns (in column 0) which are a description of an unknown plant, and an SP-grammar which includes Old SP-patterns shown in columns 1 to 6, which describe different categories of plant and a selection of their attributes. From Figure 16 in \cite{sp-extended-overview}, reproduced with permission.}
\label{class-part-plant-figure}
\end{figure}

\restoregeometry

Column 0 shows some New SP-patterns that represent features of a plant that has not yet been identified, while the SP-pattern in column 1 shows that the unknown plant is probably a Meadow Buttercup (the name is shown near the bottom of the column), and the SP-patterns in columns 2 to 6 show higher-level categories such as the genus (column 6), the family (column 5), and so on.

The way in which IC is served by Old SP-patterns like the ones shown can be seen in the way they make it possible to avoid unnecessary repetition of information. For example, the SP-pattern representing the high-level category `\texttt{Plants}' (column 2) has the features `\texttt{has\-chlorophyll}' and `\texttt{photosynthesises}'. As can be seen from the SPMA, there is no need to repeat that information in the lower-level category `\texttt{Angiospermae}' (column 3), or in the next category below, the category `\texttt{Ranunculales}' (column 4), and so on down to the level of the Meadow Buttercup (column 1).

Much the same can be said about class-inclusion hierarchies as was said about part-whole hierarchies. They promote granularity, they are very familiar, and thus likely to contribute to transparency in the results from the SP System.

\subsection{Granularity, familiarity, and SP-multiple-alignments}\label{granularity-spma-section}

As described in Section \ref{spma-icmup-section}, the concept of SPMA is a generalisation of the other six versions of ICMUP described in Section \ref{variants-of-icmup-section}. As such, it is likely to exhibit the same levels of granularity and familiarity as the other six, with corresponding benefits for transparency.

\section{Interpretability and explainability}\label{interpretability-explainability-section}

Although interpretability and explainability fall under the heading of transparency, considered in Sections \ref{transparency-via-audit-trails-section} and \ref{transparency-granularity-familiarity-section}, above, this section describes some recent studies that are more specific to those two concepts, with some brief comments from an SP perspective.

Quanshi Zhang and Song-Chun Zhu describe a survey of visual interpretability for deep learning \cite{zhang-zhu-2018}. Like other authors, they emphasise achievements with DNNs but lament how interpretability is always their Achilles' heel. Concentrating on convolutional neural networks (CNNs), they examine methods for discovering representations
of pre-trained CNNs, including methods for `disentangling' representations of pre-trained CNNs, and they examine how learning by CNNs may be achieved with `disentangled' representations, and how `middle-to-end' learning may be achieved with `model interpretability'. Finally, they suggest that ``In the future, we believe the middle-to-end learning will
continuously be a fundamental research direction.'' \cite[p.~37]{zhang-zhu-2018} In addition, they suggest that,
``based on the semantic hierarchy of an interpretable network, debugging CNN representations at the semantic level
will create new visual applications.'' ({\em ibid.}).

David Alvarez-Melis and Tommi Jaakkola describe research towards the development of neural networks that interpretable, self-explaining, and also robust \cite{alvarez-melis-jaakkola-2018}. In that connection, they propose three desirable features for neural networks: `explicitness', `faithfulness', and `stability', and they show that, in general, existing methods do not satisfy them. Starting with linear classifiers, they have developed
self-explaining models in stages, progressively generalizing them to meet their criteria of success. They say that experimental results show that the framework they have developed shows promise for reconciling the complexity of models and their interpretability.

Alejandro Barredo Arrieta and colleagues \cite{arrieta-etal-2020} present an overview of studies in ``eXplainable Artificial Intelligence (XAI)'', which they classify in two different categories: 1) ``[Machine learning] models that feature some degree of transparency, [which are] thereby interpretable to an extent by themselves; and 2) ``post-hoc XAI techniques devised to make ML models more interpretable.''(p.~108).  They have introduce a new classification of DNNs ``giving rise to an alternative taxonomy that connects more closely with the specific domains in which explainability can be realized for Deep Learning models.'' ({\em ibid.}). Also,

\begin{quote}

    ``Our reflections about the future of XAI, conveyed in the discussions held throughout this work, agree on the compelling need for a proper understanding of the potentiality and caveats opened up by XAI techniques. It is our vision that model interpretability must be addressed jointly with requirements and constraints related to data privacy, model confidentiality, fairness and accountability. ({\em ibid.}).

\end{quote}

Ruth Byrne \cite{byrne-2019} discusses how `counterfactuals' (what would have happened if circumstances had been different) may provide evidence in support of explainable AI. In particular in this connection, she considers which kinds of counterfactual are most useful. ``...~to maximize their effectiveness, it will be useful for XAI to incorporate information from psychological experiments about the way people create and comprehend counterfactuals, for counterfactuals of different structure and content, and with various relations.'' \cite[p.~6280]{byrne-2019}.

David Gunning and colleagues \cite{gunning-etal-2019} discuss how ``...~for many critical applications in defense, medicine, finance, and law, explanations are essential for users to understand, trust, and effectively manage these new, artificially intelligent partners'' (p.~1). They describe several issues associated with explainability but do not reach conclusions.

Randy Goebel and colleages \cite{goebel-etal-2018} note the problems with explainability in the results normally produced by DNNs. They suggest that one possible way forward is to develop DNNs that can create explanations in parallel with their main processing. Another possibility is some kind of hybrid process that leverages human intelligence in conjunction with machine intelligence.

These studies are only a small fraction of activity in the areas of interpretability and explainability. The impression one gains from these studies and others is that it is likely to be a struggle to develop DNNs, or varieties thereof, which provide what is needed in terms of transparency, interpretability, and explainability.

By contrast, the SP System has clear strengths in terms of `transparency via audit trails' (Section \ref{transparency-via-audit-trails-section}), and in terms of `transparency via granularity and familiarity' (Section \ref{transparency-granularity-familiarity-section}).

\section{Conclusion}

This chapter describes how the {\em SP System}---which is the {\em SP Theory of Intelligence} and the {\em SP Computer Model}---may promote transparency and granularity in AI, and perhaps also in other areas of computing.

The SP System is introduced (Section \ref{sp-outline-section}), together with an account of the significance of IC in the representation and processing of knowledge in the SP System (Section \ref{ic-kr-pr-in-sp-system-section}). It seems that much of this IC can be seen as ``IC via the matching and unification of patterns'' (ICMUP, Section \ref{icmup-section}).

An important part of ICMUP in this context is the matching of patterns that are `discontinuous', meaning that any given pattern may be interspersed with other information (Section \ref{discontinuous-patterns-section}).

Seven variants of ICMUP are described in Section \ref{variants-of-icmup-section}. Amongst those seven variants, the most important is the concept of SP-multiple-alignment (SPMA) (Sections \ref{spma-icmup-section} and \ref{concept-of-spma-section}), a version that generalises the six other versions (Section \ref{variants-of-icmup-section}).

Another important idea associated with the SP System is the concept of ``Discovery Of Natural Structures Via Information Compression'' (`DONSVIC') (Section \ref{the-donsvic-principle-section}). As described in Section \ref{the-donsvic-principle-section}, the DONSVIC principle seems to provide a basis for the concept of granularity in AI.

There may be a case for exploring what appears to be some common ground amongst such concepts as `information granule', `SP-pattern', `information chunk', and `entity' or `object' in object-oriented programming (Section \ref{tying-things-together-section}).

The main conclusions of this chapter are:

\begin{itemize}

    \item {\em Transparency via audit trails}. For the creation of any given SPMA, the SP System provides very full information about how, via heuristic search, that SPMA has been created, with full information about all the false trails that were followed in that search (Section \ref{transparency-via-audit-trails-section}). For any given SPMA it is possible to plot an audit trail of all its ancestors.

        The fact that such an audit trail can be created confirms the existence of clear, granular structures in the system's processing. There is very full information about all the SPMAs created on the path to `good' SPMAs, and all the SPMAs created on false trails away from `good' SPMAs.

    \item {\em Transparency via granularity}. The SP Computer Model has already demonstrated the unsupervised learning of words and grammatical classes from English-like artificial language without any punctuation or spaces to mark where one words ends and the next one begins. There is clear potential for further development along these lines. There is also potential for the unsupervised learning of 3D objects.

        In general, the SP System, via the DONSVIC principle, has potential to bootstrap `natural' structures in its knowledge, and thus to bootstrap granularity in that knowledge.

    \item {\em Transparency via familiarity}. Owing to the organisation and workings of the SP System, the seven variants of ICMUP described in Section \ref{variants-of-icmup-section} will be the mainstay of how its knowledge is organised.

        In view of evidence that the same principles have a role to play in brains and nervous systems \cite{sp-compression}, the kinds of structures created by the SP System as it matures are likely to be similar to structures that people use themselves---chunking-with-codes, part-whole hierarchies, class-inclusion hieararchies, and more. Consequently, the kinds of structures created by the SP System are likely to be familiar to people, helping to make those structures relatively easy to interpret and to explain, and correspondingly transparent.

\end{itemize}

Many recent studies of interpretability and explainability (Section \ref{interpretability-explainability-section}) suggest that, in the quest for transparency in those two areas, it is likely to prove difficult to overcome fundamental weaknesses in DNNs. It may be better to make a fresh start with the SP System for development, especially since there is evidence that {\em the SP System provides a relatively promising foundation for the development of artificial general intelligence.} (Section \ref{strengths-potential-section}).

Since the SP System has potential in several areas apart from AI (Section \ref{non-ai-areas-of-application-section}), there is potential for the advantages just described to be seen in those areas as well.

\bibliographystyle{iai-bib-format}

\begin{thebibliography}{10}
\expandafter\ifx\csname urlstyle\endcsname\relax
  \providecommand{\doi}[1]{doi:\discretionary{}{}{}#1}\else
  \providecommand{\doi}{doi:\discretionary{}{}{}\begingroup
  \urlstyle{rm}\Url}\fi

\bibitem{szegedy-etal-2014}
Szegedy, C., Zaremba, W., Sutskever, I., Bruna, J., Erhan, D., Goodfellow, I.,
  and Fergus, R.
\newblock Intriguing properties of neural networks.
\newblock Technical report, Google Inc.~and others (2014).
\newblock \lowercase{a}rXiv:1312.6199v4 [cs.CV] 19 Feb 2014,
  \href{http://bit.ly/1elzRGM}{bit.ly/1elzRGM} (PDF).

\bibitem{nguyen-etal-2015}
Nguyen, A., Yosinski, J., and Clune, J.
\newblock Deep neural networks are easily fooled: high confidence predictions
  for unrecognizable images.
\newblock In \emph{Proceedings of the IEEE confernce on computer vision and
  pattern recognition ({CVPR} 2015)}, pages 427--436 (2015).
\newblock \doi{10.1109/CVPR.2015.7298640}.

\bibitem{zadeh-1997}
Zadeh, L.~A.
\newblock Toward a theory of fuzzy information granulation and its centrality
  in human reasoning and fuzzy logic.
\newblock Fuzzy Sets and Systems 90, 111--127 (1997).

\bibitem{pedrycz-2018}
Pedrycz, W.
\newblock Granular computing for data analytics: a manifesto of human-centric
  computing.
\newblock {IEEE}/{CAA} Journal Of Automatica Sinica 5(6), 1025--1034 (2018).

\bibitem{sp-extended-overview}
Wolff, J.~G.
\newblock The {SP} {T}heory of {I}ntelligence: an overview.
\newblock Information 4(3), 283--341 (2013).
\newblock \doi{10.3390/info4030283}.
\newblock \lowercase{ar}Xiv:1306.3888 [cs.AI],
  \href{http://bit.ly/1NOMJ6l}{bit.ly/1NOMJ6l}.

\bibitem{wolff-2006}
Wolff, J.~G.
\newblock \emph{Unifying Computing and Cognition: the {SP} Theory and Its
  Applications}.
\newblock CognitionResearch.org, Menai Bridge (2006).
\newblock {ISBN}s: 0-9550726-0-3 (ebook edition), 0-9550726-1-1 (print
  edition). Distributors, including Amazon.com, are detailed on
  \href{http://bit.ly/WmB1rs}{bit.ly/WmB1rs}.

\bibitem{kuhn-2012}
Kuhn, T.~S.
\newblock \emph{The Structure of Scientific Revolutions}.
\newblock University of Chicago Press, Chicago and London, fourth, {K}indle
  edition (2012).

\bibitem{attneave-1954}
Attneave, F.
\newblock Some informational aspects of visual perception.
\newblock Psychological Review 61, 183--193 (1954).

\bibitem{barlow-1959}
Barlow, H.~B.
\newblock Sensory mechanisms, the reduction of redundancy, and intelligence.
\newblock In {HMSO}, editor, \emph{The Mechanisation of Thought Processes},
  pages 535--559. Her Majesty's Stationery Office, London (1959).

\bibitem{barlow-1969}
Barlow, H.~B.
\newblock Trigger features, adaptation and economy of impulses.
\newblock In Leibovic, K.~N., editor, \emph{Information Processes in the
  Nervous System}, pages 209--230. Springer, New York (1969).

\bibitem{sp-compression}
Wolff, J.~G.
\newblock Information compression as a unifying principle in human learning,
  perception, and cognition.
\newblock Complexity 2019, 38 pages (2019).
\newblock \doi{10.1155/2019/1879746}.
\newblock Article ID 1879746. viXra:1707.0161v3, hal-01624595 v2.

\bibitem{spneural-2016}
Wolff, J.~G.
\newblock Information compression, multiple alignment, and the representation
  and processing of knowledge in the brain.
\newblock Frontiers in Psychology 7, 1584 (2016).
\newblock ISSN 1664-1078.
\newblock \doi{10.3389/fpsyg.2016.01584}.
\newblock \lowercase{a}rXiv:1604.05535 [cs.AI],
  \href{http://bit.ly/2esmYyt}{bit.ly/2esmYyt}.

\bibitem{sp-micmup}
Wolff, J.~G.
\newblock Mathematics as information compression via the matching and
  unification of patterns.
\newblock Complexity 2019, 25 (2019).
\newblock \doi{10.1155/2019/6427493}.
\newblock Article ID 6427493, Archives:
  \href{http://vixra.org/abs/1912.0100}{vixra.org/abs/1912.0100} and
  \href{https://hal.archives-ouvertes.fr/hal-02395680}{hal.archives-ouvertes.fr/hal-02395680}.

\bibitem{ford-2018}
Ford, M.
\newblock \emph{Architects of Intelligence: the Truth About AI From the People
  Building It}.
\newblock Packt Publishing, Birmingham, UK, {K}indle edition (2018).

\bibitem{sp-archai-v4}
Wolff, J.~G.
\newblock Problems in {AI} research and how the {SP} {S}ystem may help to solve
  them  (2020).
\newblock Download: \href{https://tinyurl.com/y48m84t5}{tinyurl.com/y48m84t5},
  submitted for publication.

\bibitem{sp-big-data}
Wolff, J.~G.
\newblock Big data and the {SP} {T}heory of {I}ntelligence.
\newblock IEEE Access 2, 301--315 (2014).
\newblock \doi{10.1109/ACCESS.2014.2315297}.
\newblock \lowercase{ar}Xiv:1306.3890 [cs.DB],
  \href{http://bit.ly/2qfSR3G}{bit.ly/2qfSR3G}. This paper, with minor
  revisions, is reproduced in Fei Hu (Ed.), {\em Big Data: Storage, Sharing,
  and Security}, Taylor \& Francis LLC, CRC Press, 2016, Chapter 6,
  pp.~143--170.

\bibitem{sp-vision}
Wolff, J.~G.
\newblock Application of the {SP} {T}heory of {I}ntelligence to the
  understanding of natural vision and the development of computer vision.
\newblock SpringerPlus 3(1), 552--570 (2014).
\newblock \doi{10.1186/2193-1801-3-552}.
\newblock \lowercase{ar}Xiv:1303.2071 [cs.CV],
  \href{http://bit.ly/2oIpZB6}{bit.ly/2oIpZB6}.

\bibitem{wolff-sp-intelligent-database}
Wolff, J.~G.
\newblock Towards an intelligent database system founded on the {SP} theory of
  computing and cognition.
\newblock Data \& Knowledge Engineering 60, 596--624 (2007).
\newblock \doi{10.1016/j.datak.2006.04.003}.
\newblock \lowercase{a}rXiv:cs/0311031 [cs.DB],
  \href{http://bit.ly/1CUldR6}{bit.ly/1CUldR6}.

\bibitem{wolff-medical-diagnosis}
Wolff, J.~G.
\newblock Medical diagnosis as pattern recognition in a framework of
  information compression by multiple alignment, unification and search.
\newblock Decision Support Systems 42, 608--625 (2006).
\newblock \doi{10.1016/j.dss.2005.02.005}.
\newblock \lowercase{a}rXiv:1409.8053 [cs.AI],
  \href{http://bit.ly/1F366o7}{bit.ly/1F366o7}.

\bibitem{sp-palade-wolff}
Palade, V. and Wolff, J.~G.
\newblock A roadmap for the development of the `{SP} {M}achine' for artificial
  intelligence.
\newblock The Computer Journal 62, 1584--1604 (2019).
\newblock \doi{10.1093/comjnl/bxy126}.
\newblock \lowercase{h}ttps://doi.org/10.1093/comjnl/bxy126, arXiv:1707.00614,
  \href{http://bit.ly/2tWb88M}{bit.ly/2tWb88M}.

\bibitem{wolff-1988}
Wolff, J.~G.
\newblock Learning syntax and meanings through optimization and distributional
  analysis.
\newblock In Levy, Y., Schlesinger, I.~M., and Braine, M. D.~S., editors,
  \emph{Categories and Processes in Language Acquisition}, pages 179--215.
  Lawrence Erlbaum, Hillsdale, NJ (1988).
\newblock \href{http://bit.ly/ZIGjyc}{bit.ly/ZIGjyc}.

\bibitem{lenneberg-1962}
Lenneberg, E.~H.
\newblock Understanding language without the ability to speak: a case report.
\newblock Journal of Abnormal and Social Psychology 65, 419--425 (1962).

\bibitem{brown-2014}
Brown, C.
\newblock \emph{My Left Foot}.
\newblock Vintage Digital, London, {K}indle edition (2014).
\newblock First published in 1954.

\bibitem{chater-vitanyi-2007}
Chater, N. and Vit{\'a}nyi, P.
\newblock `{I}deal learning' of natural language: positive results about
  learning from positive evidence.
\newblock Journal of Mathematical Psychology 51(3), 135--163 (2007).
\newblock \doi{10.1016/j.jmp.2006.10.002}.

\bibitem{miller-1956}
Miller, G.~A.
\newblock The magical number seven, plus or minus two: some limits on our
  capacity for processing information.
\newblock Psychological Review 63, 81--97 (1956).

\bibitem{fujita-etal-2018}
Fujita, H., Gaeta, A., Loia, V., and Orciuoli, F.
\newblock Resilience analysis of critical infrastructures: a cognitive approach
  based on granular computing.
\newblock IEEE Transactions on Cybernetics 49(5), 1835--1848 (2018).
\newblock \doi{10.1109/TCYB.2018.2815178}.

\bibitem{birtwistle-etal-1973}
Birtwistle, G.~M., Dahl, O.-J., Myhrhaug, B., and Nygaard, K.
\newblock \emph{Simula Begin}.
\newblock Studentlitteratur, Lund (1973).

\bibitem{lee-liou-1996}
Lee, S.-Y. and Liou, R.-L.
\newblock A multi-granularity locking model for concurrency control in
  object-oriented database systems.
\newblock IEEE Transactions on Knowledge and Data Engineering 8(1), 144--156
  (1996).

\bibitem{sp-software-engineering}
Wolff, J.~G.
\newblock Software engineering and the {SP} {T}heory of {I}ntelligence.
\newblock Technical report, CognitionResearch.org (2017).
\newblock Submitted for publication. arXiv:1708.06665 [cs.SE],
  \href{http://bit.ly/2w99Wzq}{bit.ly/2w99Wzq}.

\bibitem{zhang-zhu-2018}
Zhang, Q. and Zhu, S.-C.
\newblock Visual interpretability for deep learning: a survey.
\newblock Frontiers of Information Technology \& Electronic Engineering 19,
  27--39 (2018).

\bibitem{alvarez-melis-jaakkola-2018}
Alvarez-Melis, D. and Jaakkola, T.~S.
\newblock Towards robust interpretabilitywith self-explaining neural networks.
\newblock In \emph{Proceedings of the 32nd Conference on Neural Information
  Processing Systems (NeurIPS 2018), Montr{\'e}al, Canada} (2018).

\bibitem{arrieta-etal-2020}
Arrieta, A.~B., D{\'i}az-Rodr{\'i}guez, N., Ser, J.~D., Bennetot, A., Tabik,
  S., Barbado, A., Garcia, S., Gil-Lopez, S., Molina, D., Benjamins, R.,
  Chatila, R., and Herrera, F.
\newblock Explainable {A}rtificial {I}ntelligence ({XAI}): {C}oncepts,
  taxonomies, opportunities and challenges toward responsible {AI}.
\newblock Information Fusion 58, 82--115 (2020).

\bibitem{byrne-2019}
Byrne, R. M.~J.
\newblock Counterfactuals in explainable artificial intelligence ({XAI}):
  evidence from human reasoning.
\newblock In \emph{Proceedings of the Twenty-Eighth International Joint
  Conference on Artificial Intelligence (IJCAI-19}, pages 6276--6282 (2019).

\bibitem{gunning-etal-2019}
Gunning, D., Stefik, M., Choi, J., Miller, T., Stumpf, S., and Yang, G.-Z.
\newblock {XAI}---{E}xplainable {A}rtificial {I}ntelligence.
\newblock Science Robotics 4(37), eaay7120 (2019).
\newblock \doi{10.1126/scirobotics.aay7120}.

\bibitem{goebel-etal-2018}
Goebel, R., Chander, A., Holzinger, K., Lecue, F., Akata, Z., Stumpf, S.,
  Kieseberg, P., and Holzinger, A.
\newblock {E}xplainable {AI}: {T}he {N}ew 42?
\newblock In \emph{CD-MAKE 2018, 27-30 Aug 2018, Hamburg, Germany, Lecture
  Notes in Computer Science, volume 11015}, pages 295--303 (2018).

\end{thebibliography}

\end{document}